\def\year{2022}\relax
\documentclass[letterpaper]{article} 
\usepackage{aaai22}  
\usepackage{times}  
\usepackage{amsmath}
\usepackage{amssymb}
\usepackage{array}
\usepackage{multirow}
\usepackage{latexsym}
\usepackage{booktabs}
\usepackage{graphicx} 
\usepackage{subfigure}
\usepackage{epstopdf}
\usepackage{comment}
\usepackage{color,xcolor}

\usepackage[hyphens]{url}  
\usepackage{float}
\usepackage{microtype}
\usepackage{makecell}
\usepackage{helvet} 
\usepackage{courier}  

\usepackage{natbib}  
\usepackage{caption} 
\title{DKPLM: Decomposable Knowledge-Enhanced Pre-trained Language Model for Natural Language Understanding}

\author{Taolin Zhang$^{1,3}$\thanks{\ \ T. Zhang and C. Wang contributed equally to this work.}, Chengyu Wang$^{2}$\footnotemark[1], Nan Hu$^{5}$, Minghui Qiu$^{2}$\thanks{\ \ Co-corresponding authors.}, Chengguang Tang$^{2}$ \\ \textbf{Xiaofeng He}$^{4,5}$\footnotemark[2], Jun Huang$^{2}$}
\affiliations {
$^1$ School of Software Engineering, East China Normal University
$^2$ Alibaba Group\\
$^3$ Shanghai Key Laboratory of Trsustworthy Computing \\
$^4$ NPPA Key Laboratory of Publishing Integration Development, ECNUP\\
$^5$ School of Computer Science and Technology, East China Normal University\\
 {zhangtl0519@gmail.com, hunan.vinny1997@gmail.com, hexf@cs.ecnu.edu.cn} \\
 {\{chengyu.wcy, minghui.qmh, chengguang.tcg, huangjun.hj\}@alibaba-inc.com}
 }

\begin{document}
\maketitle

\begin{abstract}
Knowledge-Enhanced Pre-trained Language Models (KEPLMs) are pre-trained models with relation triples injecting from knowledge graphs to improve language understanding abilities.Experiments show that our model outperforms other KEPLMs significantly over zero-shot knowledge probing tasks and multiple knowledge-aware language understanding tasks. To guarantee effective knowledge injection, previous studies integrate models with knowledge encoders for representing knowledge retrieved from knowledge graphs. The operations for knowledge retrieval and encoding bring significant computational burdens, restricting the usage of such models in real-world applications that require high inference speed. In this paper, we propose a novel KEPLM named DKPLM that
 decomposes knowledge injection process of the pre-trained language models in pre-training, fine-tuning and inference stages, which facilitates the applications of KEPLMs in real-world scenarios. Specifically, we first detect knowledge-aware long-tail entities as the target for knowledge injection, enhancing the KEPLMs' semantic understanding abilities and avoiding injecting redundant information.
 The embeddings of long-tail entities are replaced by ``pseudo token representations'' formed by relevant knowledge triples. We further design the relational knowledge decoding task for pre-training to force the models to truly understand the injected knowledge by relation triple reconstruction. Experiments show that our model outperforms other KEPLMs significantly over zero-shot knowledge probing tasks and multiple knowledge-aware language understanding tasks. We further show that DKPLM has a higher inference speed than other competing models due to the decomposing mechanism.
\end{abstract}

\section{Introduction}
Recently, Pre-trained Language Models (PLMs) improve various downstream NLP tasks significantly \cite{DBLP:conf/emnlp/HeZZCC20, DBLP:conf/acl/XuGTSSGZQJD20, DBLP:conf/acl/ChangXXT20}.
In PLMs, the two-stage strategy (i.e., pre-training and fine-tuning) \cite{DBLP:conf/naacl/DevlinCLT19} inherits the knowledge learned during pre-training and applies it to downstream tasks.
Although PLMs have stored a lot of internal knowledge, it can hardly understand external background knowledge such as factual and commonsense knowledge \cite{DBLP:journals/corr/abs-2101-12294, DBLP:conf/acl/CuiCWZ21}.
Hence, the performance of PLMs can be improved by injecting external knowledge triples, which are referred to as Knowledge-Enhanced PLMs (KEPLMs).


In the literature, the approaches of injecting knowledge can be divided into two categories, including knowledge embedding and joint learning.
(1) Knowledge embedding based approaches inject triple representations in Knowledge Graphs (KGs) trained by knowledge embedding algorithms (e.g., TransE \cite{DBLP:conf/nips/BordesUGWY13}) into contextual representations via well-designed feature fusion modules, which may contain a large number of parameters~\cite{DBLP:conf/acl/ZhangHLJSL19, DBLP:conf/emnlp/PetersNLSJSS19, su2020contextual}.
As reported in \cite{DBLP:journals/corr/abs-1911-06136}, different knowledge representation algorithms significantly impact the performance of PLMs.
(2) Joint learning based approaches learn knowledge embeddings from KGs jointly during pre-training \cite{DBLP:journals/corr/abs-1911-06136, DBLP:conf/coling/SunSQGHHZ20, DBLP:conf/aaai/LiuZ0WJD020}, which are two significantly different tasks.
We also observe that there are two potential drawbacks of previous methods.
(1) These models inject knowledge indiscriminately into all entities in pre-training sentences, which introduces redundant and irrelevant information to PLMs \cite{zhang2021drop}.
(2) Large-scale KGs are required during both fine-tuning and inference for obtaining outputs of knowledge encoders. This incurs additional computation burden that limits their usage for real-world applications that require high inference speed~\cite{DBLP:conf/emnlp/ZhangDLCZC20, DBLP:conf/acl/MalikSNGGBM20}.

\begin{figure*}
\centering
\includegraphics[height=4.5cm, width=15cm]{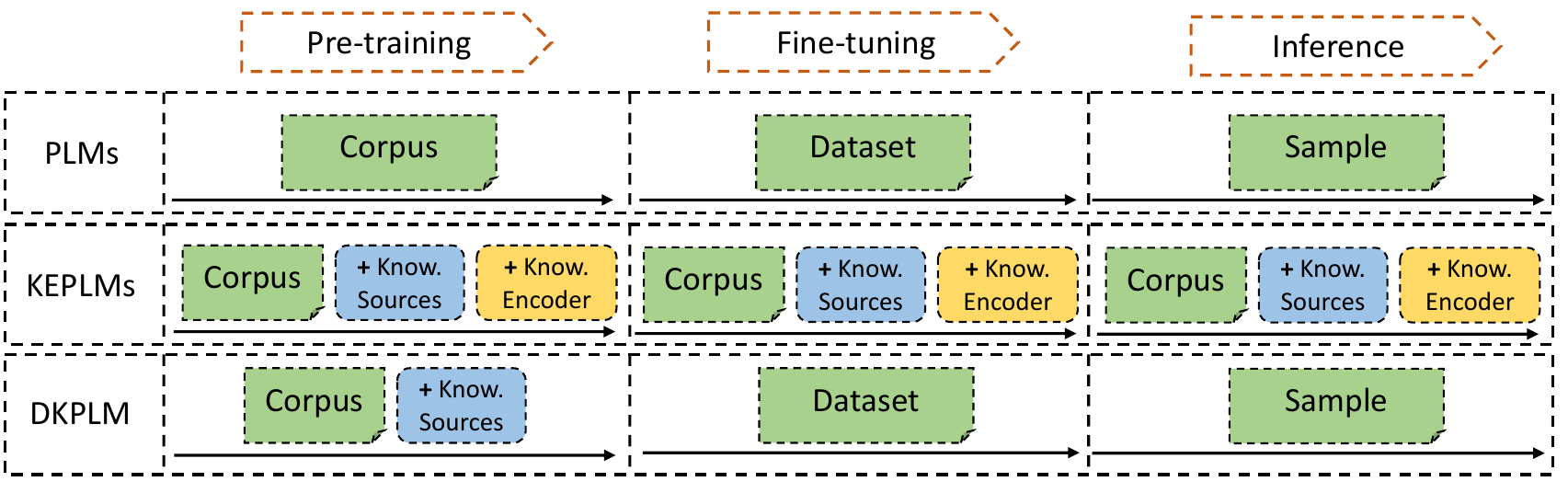}
\caption{Comparison between DKPLM and other models. (1) Plain PLMs do not utilize external knowledge in all the three stages. (2) Existing KEPLMs utilize various knowledge sources (e.g., KGs and dictionaries) to enhance understanding abilities by using knowledge encoders in all the three stages. (3) During pre-training, DKPLM utilizes the same data sources as KEPLMs, with no knowledge encoder (e.g., Neural Networks and Graph Neural Networks) required. During fine-tuning and inference, our model does not require KGs and is highly flexible and efficient. (Best viewed in color.)}
\label{motivation_figure}
\end{figure*}

To overcome the above problems, we present a novel KEPLM named DKPLM that decomposes knowledge injection process of three stages for KEPLMs. A comparison between our model and other models is shown in Figure ~\ref{motivation_figure}.
Clearly, for DKPLM, knowledge injection is only applied during pre-training, without using additional knowledge encoders. Hence, during the fine-tuning and inference stages, our model can be utilized in the same way as that of BERT \cite{DBLP:conf/naacl/DevlinCLT19} and other plain PLMs, which facilitates the applications of our KEPLM in real-world scenarios.
Specifically, we introduce three novel techniques for pre-training DKPLM:
\begin{itemize}
    \item \textit{Knowledge-aware Long-tail Entity Detection}: our model detects long-tail entities for knowledge injection based on the frequencies in the corpus, the number of adjacent entities in the KG and the semantic importance.
    In this way, we avoid learning too much redundant and irrelevant information \cite{zhang2021drop}.
    
    \item \textit{Pseudo Token Representation Injection}: we replace the embeddings of detected long-tail entities with the representations of the corresponding knowledge triples generated by the shared PLM encoder, referred to ``pseudo token representations''. Hence, the knowledge is injected without introducing any extra parameters to the model.

    
    \item \textit{Relational Knowledge Decoding}: for a relation triple, we use the representations of one entity and the relation predicate to decode each token of another entity.
    This pre-training task acts as a supervised signal for the KEPLM, forcing the model to understand what knowledge is injected to the KEPLM.
\end{itemize}

In the experiments, we evaluate our model against strong baseline KEPLMs pre-trained using the same data sources over various knowledge-related tasks, including knowledge probing (LAMA) \cite{DBLP:conf/emnlp/PetroniRRLBWM19}, relation extraction and entity typing.
For knowledge probing, the top-1 accuracy of four datasets is increased by +1.57\% on average, compared with state-of-the-art.
Meanwhile, in other tasks, our model also achieves consistent improvement. 
In summary, we make the following contributions in this paper:
\begin{itemize}
    \item We present a novel KEPLM named DKPLM to inject the knowledge into PLMs, which specifically focuses on long-tail entities, decomposing the knowledge injection process of three PLMs' stages.
    \item A dual knowledge injection process including encoding and decoding for long-tail entities is proposed to pre-train DKPLM, consisting of three modules: Knowledge-aware Long-tail Entity Detection, Pseudo Token Embedding Injection, and Relational Knowledge Decoding.
    \item In the experiments, we evaluate DKPLM over multiple public benchmark datasets, including knowledge probing (LAMA) and knowledge-aware tasks.
    Experimental results show that DKPLM consistently outperforms state-of-the-art methods.
    An analysis on inference speed of KEPLMs is also provided.
\end{itemize}

\section{Related Work}
In this section, we briefly summarize the related work on the following two aspects: PLMs and KEPLMs.

\noindent\textbf{PLMs.}
Following BERT ~\cite{DBLP:conf/naacl/DevlinCLT19}, many PLMs have been proposed to further improve performance in various NLP tasks.
We summarize the recent studies, specifically focusing on three techniques,
including self-supervised pre-training, model architectures and multi-task learning.
To improve the model's semantic understanding, several approaches extend BERT by employing novel token-level and sentence-level pre-training tasks. Notable PLMs include Baidu-ERNIE \cite{DBLP:journals/corr/abs-1904-09223}, StructBERT \cite{DBLP:conf/iclr/0225BYWXBPS20} and spanBERT \cite{DBLP:journals/tacl/JoshiCLWZL20}.
Other models boost the performance by changing the internal encoder architectures. For example,  XLNet \cite{DBLP:conf/nips/YangDYCSL19} utilizes Transformer-XL \cite{dai2019transformer} to encoder long sequences by permutation in language tokens.
Sparse self-attention \cite{cui2019fine} replaces the self-attention mechanism with more interpretable attention units.
Yet other PLMs such as MT-DNN \cite{DBLP:conf/acl/LiuHCG19} combine self-supervised pre-training with supervised learning to improve the performance of various GLUE tasks \cite{DBLP:conf/iclr/WangSMHLB19}.


\begin{figure*}[t]
\centering
\includegraphics[height=13cm, width=17.5cm]{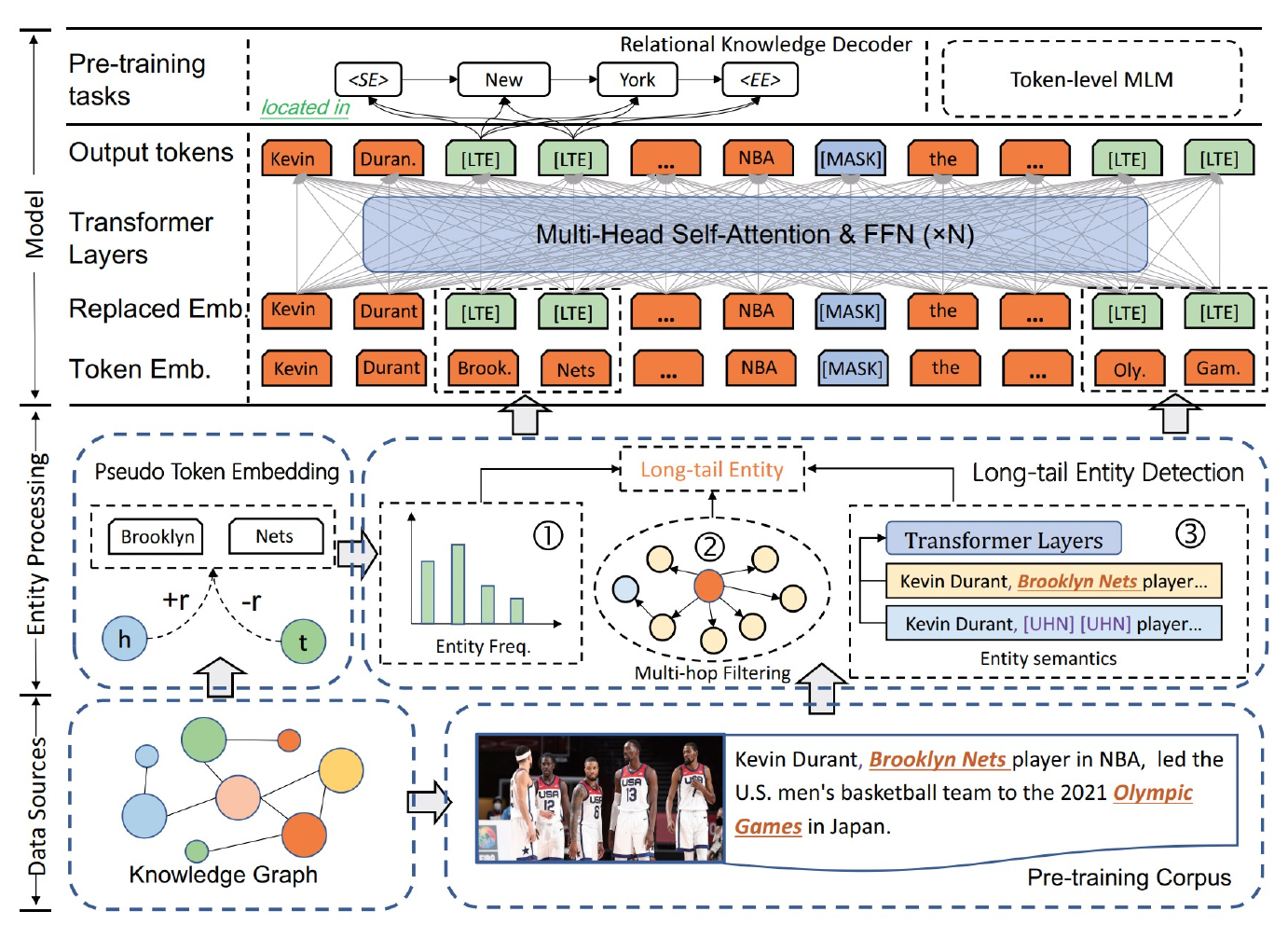}
\caption{Overview of DKPLM. (1) Data Sources: large-scale pre-training corpora and relation triplets extracted from KGs. (2) Input Data: DKPLM detects long-tail entities and retrieves relation triples for learning ``pseudo token embeddings''.
(3) Model: Plain PLMs can be used as model backbones.
We also propose the relational knowledge decoder for better knowledge injection. (Best viewed in color.)}
\label{model_figure}
\end{figure*}


\noindent\textbf{KEPLMs.}
As plain PLMs are only pre-trained on large-scale unstructured corpora, they lack the language understanding abilities of important entities.
Hence, KEPLMs use structured knowledge to enhance the language understanding abilities of PLMs.
We summary recent KEPLMs grouped into the following three types:

(1) Knowledge-enhancement by entity embeddings. For example, ERNIE-THU \citep{DBLP:conf/acl/ZhangHLJSL19} injects entity embeddings into contextual representations via knowledge-encoders stacked by the information fusion module. KnowBERT \citep{DBLP:conf/emnlp/PetersNLSJSS19} introduces the knowledge attention and recontextualization (KAR) and entity linking to inject knowledge embeddings to PLMs.
(2) Knowledge-enhancement by entity descriptions. These studies learn entity embeddings by knowledge descriptions.
For example, pre-training corpora and entity descriptions in KEPLER \citep{DBLP:journals/corr/abs-1911-06136} are encoded into a unified semantic space within the same PLM.
(3) Knowledge-enhancement by converted triplet's texts. K-BERT \citep{DBLP:conf/aaai/LiuZ0WJD020} and CoLAKE \citep{DBLP:conf/coling/SunSQGHHZ20} convert relation triplets into texts and insert them into training samples without using pre-trained embeddings.
These KEPLMs require knowledge encoder modules with additional parameters to inject the knowledge into context-aware hidden representations generated by PLMs. Previous studies \cite{DBLP:conf/emnlp/PetroniRRLBWM19, DBLP:conf/conll/Broscheit19, DBLP:journals/corr/abs-2010-11967, DBLP:conf/acl/CaoLHSYLXX20} have also shown that the semantics of high-frequency and general knowledge triples are already captured by plain PLMs, and express redundant knowledge. In this paper, we argue that enhancing the understanding ability of long-tail entities can further benefit the context-aware representations of PLMs, which is one of the major focus of this work.



\section{DKPLM: The Proposed Model}

We first state some basic notations.
Denote an input sequence of tokens as $\{w_1, w_2, \ldots, w_n\}$, where $n$ is the length of the input sequence.
The hidden representation of input tokens obtained by PLMs is denoted as $\{h_1, h_2, \ldots, h_n\}$ and $h_i \in \mathbb{R}^{d_1}$, where $d_1$ is the dimension of the PLM's output.
Furthermore, we denote the knowledge graph as $\mathcal{G}=(\mathcal{E}, \mathcal{R})$ where $\mathcal{E}$ and $\mathcal{R}$ are the collections of entities and relation triples, respectively.
In the KG, a relational knowledge triple is denoted as $(e_h, r, e_t)$, where $e_h$ and $e_t$ refer to the head entity and the tail entity, respectively.
$r$ is the specific relation predicate between $e_h$ and $e_t$.


Our pre-training process specifically focuses on the knowledge injection and decoding on certain tokens in the detected long-tail entities. Specifically, we aim to solve three research questions:
\begin{itemize}
    \item \textbf{RQ1}: What types of tokens in the pre-training corpus should be detected for knowledge injection ?
    \item \textbf{RQ2}: How can we inject knowledge to selected tokens without additional knowledge encoders ?
    \item \textbf{RQ3}: How can we verify the effectiveness of injected relation triples during pre-training ?
\end{itemize}

The overall framework of DKPLM is presented in Figure \ref{model_figure}. 
In the following, we introduce the techniques of DKPLM and discuss how we can address the three research questions.

\subsection{Knowledge-aware Long-tail Entity Detection}

\begin{figure}
\centering
\includegraphics[height=5cm, width=7cm]{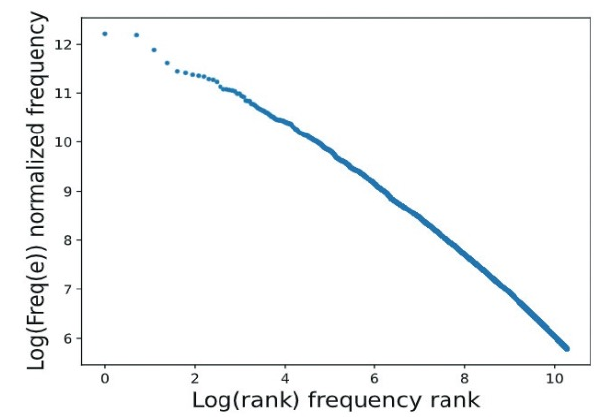}
\caption{The distribution of entity frequencies in the English Wikipedia corpus.}
\label{zipf_law}
\end{figure}

\noindent{\bf Motivation and Analysis.}
We first extract structured knowledge triples from large-scale KGs, and link entities in KGS to the target mentions in the pre-training samples by entity linking tools (e.g., TAGME \cite{DBLP:conf/cikm/FerraginaS10}).
For a better understanding of how entities are distributed in the corpus, 
we plot the distribution of the entity frequencies in the entire Wikipedia corpus, shown in Figure \ref{zipf_law}. As seen, it closely follows the~\emph{power-law distribution} with the formula as follows:
\begin{equation}
Freq(e) = \frac{C}{rank^\alpha},
\end{equation}
where $C$ and $\alpha$ are hyper-parameters, $rank$ is the entity frequency rank and $Freq(e)$ is the frequency of the entity $e$.
We can see that, while a few entities frequently appear, most of the entities seldom occur in the pre-training corpus, making it difficult for PLMs to learn better contextual representations.

As reported by~\cite{zhang2021drop}, the high-frequency relational triples are injected into PLMs is NOT always beneficial for downstream tasks. This practice is more likely to trigger negative knowledge infusion. 
Hence, knowledge injection for long-tail entities instead of all entities that occur in the corpus may further improve the understanding abilities of PLMs.
It should be further noted that the above analysis of entities only considers the frequencies in the pre-training corpus, ignoring the information of each entity in KGs and the importance of such entities in a sentence. 
In the following, we present the~\emph{Knowledge-aware Long-tail Entity Detection} mechanism to select target entities for knowledge injection.

\noindent{\bf Our Approach.} 
In our work, we consider three neighboring information of entities to characterize the ``long-tailness'' property of entities, namely the entire pre-training corpus, current input sentence and the KG. For a specific entity $e$, we consider the three following factors:
\begin{itemize}
    \item \textbf{Entity Frequency}: the entity frequency w.r.t. the entire pre-training corpus, denoted as $Freq(e)$;
    
    \item \textbf{Semantic Importance}:  the importance of the entity $e$ in the sentence, denoted as $SI(e)$;
    
     \item \textbf{Knowledge Connectivity}: the number of multi-hops neighboring nodes w.r.t. the entity $e$ in the KG, denoted as $KC(e)$.
\end{itemize}

While the computation of $Freq(e)$ is quite straightforward, it is necessary to elaborate the computation of $SI(e)$ and $KC(e)$.
$SI(e)$ refers to the semantic similarity between the representation of the sentence containing the entity $e$ and the representation of the sentence with $e$ being replaced. The greater the similarity between sentences is, the smaller the influence on the sentence semantics is when the entity is replaced. Denote $h_o$ and $h_{rep}$ as the representations of the original sentence and the sentence after entity replacing.
For simplicity, we use the reciprocal of cosine similarity to measure $SI(e)$:
\begin{equation}
    SI(e) = \frac{\left \| h_o^{T} \right \|\cdot \left \| h_{rep}\right \|}{h_o^{T}\cdot h_{rep}}
\end{equation}
In the implementation, we use the special token ``[UHN]'' to replace the entity $e$ in the sentence.

$KC(e)$ represents the importance of entity $e$ in triples' neighboring structure and we use the multi-hops neighboring entities' number to calculate $KC(e)$:
\begin{gather}
     KC(e) = \left[\left|\mathcal{N}\left(e\right)\right|\right]_{\mathrm{R}_{\min }}^{\mathrm{R}_{\max }} \\
    \mathcal{N}(e) \triangleq\left\{e^{\prime} \mid \mathrm{Hop}\left(e^{\prime}, e\right)<\mathrm{R}_{\mathrm{hop}} \wedge e^{\prime} \in \mathcal{E}\right\}
\end{gather}
where $R_{min}$ and $R_{max}$ are pre-defined thresholds. Specifically, we constrain the number of hops for computing $KC(e)$ is in the range of $R_{min}$ to $R_{max}$.
$|\cdot|$ means the neighboring entity number in the set.
The $\mathrm{Hop}$ function denotes the number of multi-hops between entity $e$ and entity $e^{\prime}$ in KGs' structure.
The degree of the ``knowledge-aware long-tailness'' $KLT(e)$ of the entity $e$ is then calculated as:
\begin{equation}
KLT(e)=\mathbb{I}_{\left\{ Freq(e)<\mathrm{R}_{freq}\right\}}\cdot SI(e)\cdot KC(e)
\end{equation}
where the $\mathbb{I}_{\{x\}}$ is the indicator function with $x$ to be a Boolean expression.
$R_{freq}$ is a pre-defined threshold.
Given a sentence, we detect all the entities and regard the entities with the $KLT(e)$ score lower than the average as knowledge-aware long-tail entities.

\subsection{Pseudo Token Embedding Injection}

In order to enhance the PLMs' understanding abilities of long-tail entities, we inject knowledge triples into the positions of such entities without introducing any other parameters.
Inspired by the KG embedding algorithms~\cite{DBLP:conf/nips/BordesUGWY13}, if an entity in the pre-training sentence is a head entity $e_h$ of knowledge triples, the representation of $e_h$ is modeled by the following function:
\begin{equation}
h_{e_h} = h_{e_t} - h_{r}
\end{equation}
where $h_{e_h}$, $h_{e_t}$ and $h_{r}$ are the representations of the head entity $e_h$, the tail entity $e_t$ and the relation predicate $r$, respectively.
Similarly, if an entity is a tail entity $e_t$ in the KG, we have:
$h_{e_t} = h_{e_h} + h_{r}$.

Specifically, we use the underlying PLM as the shared encoder to acquire the knowledge representations. Consider the situation where the entity is a head entity $e_h$ in the KG. We concatenate the tokens of the tail entity $e_t$ and the relation predicate $r$, and feed them to the PLM. The token representations of the last layer of the PLM are denoted as $\mathcal{F}(e_t)$ and $\mathcal{F}(r)$, respectively.

The pseudo token representations $h_{e_t}$ and $h_{r}$ are then computed as follows:
\begin{gather}
h_{e_t}=\mathcal{LN}\left( \sigma \left(f_{sp}\left(\mathcal{F}(e_t)\right)W_{e_t}\right)\right)\\
h_{r}=\mathcal{LN}\left( \sigma \left(f_{sp}\left(\mathcal{F}(r)\right)W_{r}\right)\right)
\end{gather}
where $\mathcal{LN}$ is the LayerNorm function \cite{DBLP:journals/corr/BaKH16} and $f_{sp}$ is the self-attentive pooling operator \citep{DBLP:conf/iclr/LinFSYXZB17} to generate the span representations.
$W_{e_t}$ and $W_{r}$ are trainable parameters.

Since the lengths of the entity and the relation predicate are usually short, the generated representations by the PLM may be not expressive.
We further consider the description text of the target entity, denoted as  $e_{h}^{des}$. Let $\mathcal{F}(e_{h}^{des})$ be the token sequence representations of $e_{h}^{des}$, generated by the PLM. We denote the pseudo token embedding $h_{e_h}$ of the head entity $e_h$ as follows:
\begin{gather}
h_{e_h} = \tanh \big((h_{e_t}-h_{r}) \oplus \mathcal{F}(e_{h}^{des}) \big)W_{e_h},
\end{gather}
where $\oplus$ refers to the concatenation of two representations, and $W_{e_h}$ is the trainable parameter.


Finally, we replace the representations of detected long-tail entities with the pseudo token representations in the PLM's embedding layer (either $h_{e_h}$ or $h_{e_t}$, depending on whether the target entity is the head or the tail entity in the KG). This follows the successive multiple transformer encoder layers to incorporate the knowledge into the contextual representations without introducing any other new parameters for knowledge encoding.

\begin{table*}[t]
\centering
\begin{small}
\begin{tabular}{c|ccc|ccc|cc}
\toprule
 \multirow{2}*{Datasets} & \multicolumn{3}{c}{PLMs} & \multicolumn{5}{c}{KEPLMs} \\
\cmidrule(r){2-4} \cmidrule(r){5-9}
~ & ELMo & BERT & RoBERTa &  CoLAKE & K-Adapter $^\ast$ & KEPLER & DKPLM & $\bigtriangleup$ 	 \\
\midrule
Google-RE & 2.2\% & 11.4\% & 5.3\%  & 9.5\%  & 7.0\% & 7.3\% & \textbf{10.8}\% & +1.3\% \\
UHN-Google-RE & 2.3\% & 5.7\% & 2.2\% & 4.9\%  & 3.7\% & 4.1\% & \textbf{5.4}\% & +0.5\%  \\ \midrule
T-REx & 0.2\%  & 32.5\% & 24.7\% & 28.8\%  & 29.1\% & 24.6\% & \textbf{32.0}\% & +2.9\%  \\
UHN-T-REx & 0.2\% & 23.3\% & 17.0\% & 20.4\% & \textbf{23.0}\%  & 17.1\% & 22.9\% & -0.1\% \\ \bottomrule
\end{tabular}
\end{small}
\caption{The performance on knowledge probing datasets. $\bigtriangleup$ represents an improvement over the best results of existing KEPLMs compared to our model.
Besides, K-Adapter $^\ast$ is based on RoBERTa-large and uses a subset of T-REx as its training
data, which may contribute to its superiority over the other KEPLMs and is unfair for DKPLM to be compared against.}
\label{knowledge_probing_result}
\end{table*}

\subsection{Relational Knowledge Decoding}

After the information of the relation triples has been injected into the model, it is not clear whether the model has understood the injected knowledge.
We design a relational knowledge decoder, forcing our model to understand the injected knowledge explicitly.
Specifically, we employ a self-attention pooling mechanism to obtain the masked entity span representations in the last layer:
\begin{equation}
    h_{e_h}^{o}=\mathcal{LN}\left( \sigma \left(f_{sp}\left(\mathcal{F}(h_{e_h})\right)W_{e_h}^{o}\right)\right)
\end{equation}
where $W_{e_h}^{o}$ is the learnable parameter.
Given the output representation of the head entity $ h_{e_h}^{o}$ and the relation predicate $h_r$, we aim to decode the tail entity.~\footnote{If the target entity is the tail entity, we can also decode the head entity in a similar fashion.}
Let $h_d^{i}$ be the representation of the $i$-th token of the predicted tail entity. We have:
\begin{equation}
        h_d^{i} = \tanh(\delta_d h_{r_i} h_{d}^{i-1}\cdot W_d),
\end{equation}
where $\delta_d$ is a scaling factor, and $h_{d}^{0}$ equals to $h_{e_h}^{o}$ as the initialization heuristics.

Because the vocabulary size is relatively large, we use the Sampled SoftMax function \cite{DBLP:conf/acl/JeanCMB15} to compare the prediction results against the ground truth. The token-level loss function $\mathcal{L}_{{d}_i}$ is defined as follows:
\begin{gather}
\mathcal{L}_{{d}_i} = \frac{\exp(f_s(h_d^{i},y_i))}{\exp(f_s(h_d^{i},y_i)) + N {\mathbb{E}_{t_n\sim Q(y_n|y_i)}}[\exp (f_s(h_d^{i},y_n))]} \\
f_s(h_d^{i},y_i)) = (h_d^{i})^{T}\cdot y_i - \log \big(Q(t|t_i) \big)
\end{gather}
where $y_i$ is the ground-truth token and $y_n$ is the negative token sampled in $Q(t_n|t_i)$.
$Q(\cdot|\cdot)$ is the negative sampling function (to be described in the experiments). $N$ is the number of negative samples.
In our DKPLM model, the training objectives include two pre-training tasks: (1) relational knowledge decoding and (2) token-level Masked Language Modeling (MLM), as proposed in \cite{DBLP:conf/naacl/DevlinCLT19}.
Hence, the total loss function of DKPLM can be denoted as follows:
\begin{equation}
\mathcal{L}_{\mathrm{total}}=\lambda_1\mathcal{L}_{\mathrm{MLM}}+(1-\lambda_1)\mathcal{L}_{\mathrm{De}}
\end{equation}
where the $\lambda_1$ is the hyper-parameter and $\mathcal{L}_{\mathrm{De}}$ is the total decoder loss of the target entity consisting of multiple tokens.

\section{Experiments}

\subsection{Pre-training Data}
In this paper, we use English Wikipedia (2020/03/01) \footnote{https://dumps.wikimedia.org/enwiki/} as our pre-training data source, and WikiExtractor \footnote{https://github.com/attardi/wikiextractor} to process the downloaded Wikipedia dump, similar to CoLAKE \cite{DBLP:conf/coling/SunSQGHHZ20} and ERNIE-THU \cite{DBLP:conf/acl/ZhangHLJSL19}.
We use Wikipedia anchors to align the entities in the pre-training texts recognized by entity linking tools (e.g., TAGME \cite{DBLP:conf/cikm/FerraginaS10}) to WikiData5M \cite{DBLP:journals/corr/abs-1911-06136}, which is a large-scale proposed KG data source including relation triples and entity description texts.
The additional pre-processing and filtration steps are kept the same as ERNIE-THU \cite{DBLP:conf/acl/ZhangHLJSL19}.
In total, we have 
3,085,345 entities and 822 relation types in the KG and 26 million training samples in our pre-training corpus.


\subsection{Baselines}
We consider the following models as strong baselines:

\noindent{\textbf{ERNIE-THU}} \cite{DBLP:conf/acl/ZhangHLJSL19}: The model integrates a denoising entity auto-encoder pre-training task to inject knowledge embeddings into language representations.
\noindent{\textbf{KnowBERT}} \cite{DBLP:conf/emnlp/PetersNLSJSS19}: 
It injects rich structured knowledge representations via knowledge attention and recontextualization.
\noindent{\textbf{KEPLER}} \cite{DBLP:journals/corr/abs-1911-06136}: The model encodes texts and entities into a unified semantic space with the same PLM as the shared encoder.
\noindent{\textbf{CoLAKE}} \cite{DBLP:conf/coling/SunSQGHHZ20}: It considers a unified heterogeneous KG as the knowledge source and employs adjacency matrices to control the information flow.
\noindent{\textbf{K-Adapter}} \cite{DBLP:conf/acl/WangTDWHJCJZ21}: 
It uses different representations for different types of knowledge via neural adapters.


\subsection{Knowledge Probing}
The knowledge probing tasks, called LAMA (LAnguage Model Analysis) \cite{DBLP:conf/emnlp/PetroniRRLBWM19}, aim to measure whether the factual knowledge is stored in PLMs via cloze-style tasks.
The LAMA-UHN tasks \cite{DBLP:journals/corr/abs-1911-03681} are proposed to alleviate the problem of overly relying on the surface form of entity names, and are constructed by filtering out the easy-to-answer samples.
These two tasks are evaluated under the zero-shot setting without fine-tuning, which is a fair comparison of the knowledge understanding abilities of KEPLMs.
We report the macro-averaged mean precision (P@1) of DKPLM.

The performance of LAMA and LAMA-UHN tasks is summarized in Table \ref{knowledge_probing_result}.
Compared to the results of other baselines, we can draw the following conclusions.
(1) BERT outperforms RoBERTa by a large gap (+5.93\% on average) because its vocabulary size is much smaller than RoBERTa. 
(2) Although our model is trained on RoBERTa-base, it achieves state-of-the-art results over three datasets (+1.57\% on average).
The result of our model is only 0.1\% lower than K-Adapter, without using any T-REx training data and large PLM backbone.
From the overall results, we can see that our learning process based on long-tail entities can effectively store and understand factual knowledge from KGs.

\begin{table}
\centering
\begin{small}
\begin{tabular}{l | ccc}
\toprule
Model & Precision & Recall & F1 \\
\midrule
UFET \citep{DBLP:conf/acl/LevyZCC18} & 77.4 & 60.6 & 68.0 \\
BERT & 76.4 & 71.0 & 73.6 \\
RoBERTa & 77.4 & 73.6 & 75.4 \\ \midrule
ERNIE$_{BERT}$ & 78.4 &72.9 &75.6  \\
ERNIE$_{RoBERTa}$ & \textbf{80.3} &70.2 &74.9  \\
KnowBERT$_{BERT}$ & 77.9 &71.2 &74.4  \\
KnowBERT$_{RoBERTa}$ & 78.7& 72.7 &75.6  \\
KEPLER$_{WiKi}$ & 77.8& 74.6& 76.2  \\
CoLAKE & 77.0 &75.7 &76.4  \\ \midrule
DKPLM & 79.2 & \textbf{75.9} & \textbf{77.5} \\ \bottomrule
\end{tabular}
\end{small}
\caption{The performance of models on Open Entity (\%).}
\label{open_entity_result}
\end{table}

\begin{figure*}
\centering
\includegraphics[height=5.5cm, width=17cm]{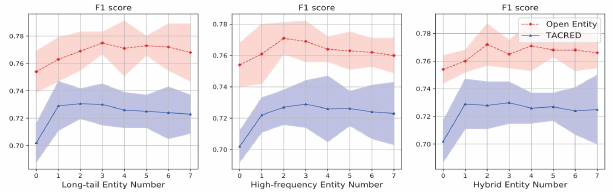}
\caption{The influence of different injected numbers of long-tail entities and high-frequency entities.}
\label{entity_type_influence}
\end{figure*}

\subsection{Knowledge-Aware Tasks}
We evaluate our DKPLM model over the knowledge-aware tasks, including relation extraction and entity typing.

\noindent\textbf{Entity Typing}: Unlike Named Entity Recognition \cite{DBLP:conf/acl/JiangZCYZ20, DBLP:conf/acl/0001MTZW020}, entity typing requires the model to predict fine-grained entity types in given contexts. 
We fine-tune our DKPLM model over Open Entity \cite{DBLP:conf/acl/LevyZCC18}.
    Table \ref{open_entity_result} shows the performance of various models including PLMs, KEPLMs and other taks-specific models.
    From the results, we can observe: the KEPLMs outperform task-specific models and the plain PLMs. 
    In addition, our DKPLM model with injected long-tail entity knowledge achieves a large performance gain compared to baselines (+2.2\% Precision, +0.2\% Recall and +1.1\% F1). 
    
\noindent\textbf{Relation Extraction}: The Relation Extraction task (RE) aims to determine the fine-grained semantic relation between the two entities in a given sentence.
We use a benchmark RE dataset TACRED \cite{DBLP:conf/emnlp/ZhangZCAM17} to evaluate our model's performance.
The relation types in TACRED is 42 and we adopt the micro averaged metrics and macro averaged metrics for evaluation.
As shown in Table \ref{tacred_result}, the performance of knowledge-injected models are much higher, and our model achieves new state-of-the-art performance (+1.46\% F1), which implies injecting long-tail entities' knowledge triple into PLMs for RE is also very effective.

\begin{table}
\centering
\begin{small}
\begin{tabular}{l|ccc}
\toprule
Model & Precision & Recall & F1 \\
\midrule
CNN & 70.30 & 54.20 & 61.20 \\
PA-LSTM \citep{DBLP:conf/emnlp/ZhangZCAM17} & 65.70  & 64.50 & 65.10 \\
C-GCN \citep{DBLP:conf/emnlp/Zhang0M18} & 69.90 & 63.30 & 66.40 \\
BERT & 67.23 & 64.81 & 66.00  \\
RoBERTa & 70.80& 69.60 & 70.20  \\ \midrule
ERNIE$_{BERT}$ & 70.01 & 66.14 & 68.09  \\
KnowBERT & 71.62 & 71.49 & 71.53 \\ \midrule
DKPLM & \textbf{72.61} & \textbf{73.53} & \textbf{73.07} \\ \bottomrule
\end{tabular}
\end{small}
\caption{The performance of models on TACRED (\%).}
\label{tacred_result}
\end{table}

\subsection{Analysis of Running Time}
In this section, we compare DKPLM with other models on pre-training, fine-tuning and inference time.
Specifically, we choose 1000 samples randomly from the pre-training corpus and two knowledge-aware tasks.
The fine-tuning and inference time is the average time of the two datasets, respectively.

As shown in Table \ref{inference_time_model_parameter}, we have the following observations.  (1) The running time of the three stages of plain PLMs are consistently shorter than existing KEPLMs due to the smaller size of model parameters. Specifically, existing KEPLMs contain the knowledge encoder module to project the knowledge embedding space to the contextual semantic space. (2) The running time of our model (especially during model fine-tuning and inference) is very similar to that of plain PLMs. The reason for the slightly longer time is that DKPLM adds a few projection parameters to align the knowledge triple representations.
This experiment shows that DKPLM is useful for online applications due to its fast inference speed.

\begin{table}
\centering
\begin{small}
\begin{tabular}{l|ccc}
\toprule
Model & Pre-training & Fine-tuning & Inference \\
\midrule
RoBETa\_{base} & 9.60 & 7.09 & 1.55 \\
BERT\_{base} & 8.46 & 6.76 & 0.97\\
\midrule
ERNIE-THU & 14.71 & 8.19 & 1.95 \\
KEPLER & 18.12 & 7.53 & 1.86\\
CoLAKE & 12.46 & 8.02 & 1.91\\ \midrule
DKPLM & 10.02 & 7.16 & 1.61\\ \bottomrule

\end{tabular}
\end{small}
\caption{The running time (s) in the three stages of various PLMs and KEPLMs over 1000 random samples.}
\label{inference_time_model_parameter}
\end{table}

\subsection{Influence of Long-tail and High-frequency Entities}
We evaluate DKPLM using different injected entity numbers, and consider three types including long-tail entities only, high-frequency entities only and a mixture of these entities.
We choose TACRED and Open Entity datasets and report the F1 metrics over testing sets to verify the effectiveness of knowledge injection.
As shown in Figure \ref{entity_type_influence}, we can observe that: (1) 
Injecting knowledge triples into long-tail entities is better than high-frequency entities. (2) The state-of-the-art performance can be obtained by injecting knowledge to a fewer entities rather than all the entities.
(3) 
Our results are consistent with \citet{zhang2021drop} in that injecting too much knowledge may hurt the performance.

\begin{table}
\centering
\begin{small}
\begin{tabular}{l cc}
\toprule
Model & TACRED & Open Entity \\\midrule
DKPLM & \bf 73.07\% & \bf 77.5\% \\ 
\quad \textbf{-} Long-tail Entity Detection  & 72.89\%  &  77.3\%\\
\quad \textbf{-} Pseudo Token Embedding & 72.35\% & 76.7\%  \\
\quad \textbf{-} Knowledge Decoding & 72.54\% &  77.1\%  \\ \bottomrule
\end{tabular}
\end{small}
\caption{Ablation study on two tasks (testing sets).}
\label{ablation_study}
\end{table}

\subsection{Ablation Study}
We report DKPLM's performance in two knowledge-aware testing sets to perform the ablation study on the F1 metric.
As shown in Table \ref{ablation_study}, we can conclude that (1) our proposed three mechanisms are effective in contributing to the complete DKPLM model.
(2) The model's performance declines significantly when removing the ``Pseudo Token Embedding'' mechanism. Here, the external knowledge of detected long-tail entities is not injected into the model.
DKPLM degenerates to relying entirely on entity-level information to decode knowledge triples, leading to model confusion due to the sparsity of the knowledge of long-tail entities.

\section{Conclusion and Future Work}
In this paper, we propose a novel KEPLMs to decouple knowledge injection and fine-tuning for knowledge-enhanced language understanding named DKPLM.
In DKPLM, we design three entity-related mechanisms to inject the knowledge information into the PLMs with minimum extra parameters for the real-world scenarios, namely knowledge-aware long-tail entity detection, pseudo token embedding injection and relational knowledge decoding.
The experiments show that our model achieves the state-of-the-art performance over zero-shot knowledge probing tasks and knowledge-aware downstream tasks.
Future work includes (1) selecting more effective knowledge triples from large-scale KGs to inject external knowledge into the PLMs, and (2) utilizing noised knowledge triples to further enhance the language understanding abilities of PLMs.
\section{Acknowledgements}
We would like to thank the anonymous reviewers for their valuable comments.
This work is supported by the Alibaba Group through Alibaba Research Intern Program.
\bibliography{aaai22}

\end{document}